\documentclass{article}
\usepackage{ijcai24}

\usepackage{times}
\usepackage{soul}
\usepackage{url}
\usepackage[hidelinks]{hyperref}
\usepackage[utf8]{inputenc}
\usepackage[small]{caption}
\usepackage{graphicx}
\usepackage{amsmath}
\usepackage{amsthm}
\usepackage{booktabs}
\usepackage[switch]{lineno}

\usepackage{color, soul}
\usepackage{xurl}
\usepackage{bm}
\usepackage{amssymb}

\usepackage{calrsfs}
\DeclareMathAlphabet{\pazocal}{OMS}{zplm}{m}{n}
\SetMathAlphabet\pazocal{bold}{OMS}{zplm}{bx}{n}

\usepackage{subcaption}
\usepackage{algorithm}
\usepackage[noend]{algpseudocode}
\usepackage{listings}
\usepackage{rotating}
\usepackage{multirow}

\def\sequitur{\textsc{Sequitur}}
\def\repair{\textsc{Repair}}
\def\longestFirst{\textsc{LongestFirst}}
\def\mostCompressive{\textsc{MostCompressive}}
\def\lz{\textsc{Lz78}}

\newcommand{\fn}[1]{\texttt{#1}}
\newcommand{\argmax}{\mathop{\mathrm{argmax}}\nolimits}
\algrenewcommand\algorithmicindent{0.5em}%
\usepackage{fancyhdr}
\title{Musical Phrase Segmentation via Grammatical Induction}

\author{
Reed Perkins
\And
Dan Ventura
\affiliations
Computer Science Department\\
Brigham Young University\\
\emails
reedperkins32@gmail.com,
ventura@cs.byu.edu
}

\begin{document}
\thispagestyle{fancy}%

\maketitle

\begin{abstract}
We outline a solution to the challenge of musical phrase segmentation that uses grammatical induction algorithms, a class of algorithms which infer a context-free grammar from an input sequence. We analyze the performance of five grammatical induction algorithms on three datasets using various musical viewpoint combinations. Our experiments show that the  \longestFirst{} algorithm achieves the best F1 scores across all three datasets and that input encodings that include the \texttt{duration} viewpoint result in the best performance.
\end{abstract}


\section{Introduction}\label{sec:intro}
Procedural generation exploits the fact that complex structures can be created by iteratively applying a set of transformation rules $R$ to some axiom $a$, possibly verifying the output against constraints $C$. We can describe an algorithm for procedural generation as a function $f: \pazocal{R} \times A \times \pazocal{C} \rightarrow \Omega$ that maps a set of rules $R \in \pazocal{R}$, an axiom $a \in A$, and a set of constraints $C \in \pazocal{C}$ to an output artifact $o \in \Omega$.\footnote{Often a random seed $\sigma$ is included as a parameter, but we omit this for simplicity. Additionally, $R$ may be paramaterized by a vector of probabilities $\bm{\theta}$, but for notational clarity we omit this as well.}  For example, $R$ could be rules that govern tree branch growth, $a$ a sapling, $C$ restrictions on the length of branches, and $o$ a rendered image of a tree; or, $R$ could represent path transitions to different story lines, $a$ an initial plot, $C$ detection of continuity errors, and $o$ the backstory for an NPC in a video game. 

Popular implementations of $f$ including probabilistic context-free grammars and genetic algorithms are capable of producing impressive results across a variety of domains, including visual arts and video games \cite{muller_procedural_2006,stava_inverse_2010,talton_learning_2012,ritchie_controlling_2015,ritchie_example-based_2018,rowe_inspired_2016}, narrative stories \cite{mason_lume_2019}, and musical sequences \cite{mccormack_grammar_1996,steedman_generative_1984,bod_probabilistic_2001,bryden_using_2006,cilibrasi_algorithmic_2004,dalhoum_computer-generated_2008,fox_genetic_2006,gilbert_probabilistic_2007,kitani_improvgenerator:_2010}. We are primarily concerned with the latter category --- procedural generation algorithms for symbolic musical sequences. As a thought exercise, we propose a generator $f$ that is capable of procedurally generating musical exercises.  We would to like to be able to leverage the power of existing procedural generation algorithms to generate new musical exercises, and so we must supply our musical exercise generator $f$ with a careful choice of $R$, $a$, and $C$.

Ignoring for a moment the exact details of such a system, we must ask ourselves how might $R$ be derived. In the instances of procedural generation algorithms cited above, there are two general approaches to this problem: either $R$ is manually created via experimentation or it is derived in some manner from a dataset of examples $\pazocal{D}$. The latter option is more appealing when designing an autonomous creative system since it removes one degree of hardcoded human intervention. In order for this technique to succeed, however, the dataset $\pazocal{D}$ must contain examples that are implicitly ``tokenized'' --- that is, examples that are imbued with some kind of structural metadata. This allows an induction technique $I$ to infer the set of rules $R$ from $\pazocal{D}$. For example, in \cite{talton_learning_2012} the authors use a dataset of 3D spaceship models in conjunction with Bayesian optimization and Monte Carlo sampling in order to produce $R$. The results are visually stunning but are also invariably predicated upon the fact that their dataset $\pazocal{D}$ is populated with labeled component hierarchies. This is a crucial caveat---an induction algorithm $I$ that derives $R$ from a dataset $\pazocal{D}$ of examples  presupposes that those examples are encoded as (possibly hierarchical) labeled components. Unfortunately for our purposes, it is not obvious how to apply a given induction technique $I$ to a set of musical examples because musical sequences lack explicit structural semantics. 

This is not to say, of course, that musical sequences lack structure. Indeed, music libraries are filled with publications that offer an interpretation of the structure and salient motifs of musical sequences \cite{barlow_dictionary_1953,bruhn_js_1993,schoenberg_fundamentals_1967}, and developing the skills to recognize and operate within musical structure is an integral part of musical education. Additionally, there have been many attempts to formalize the exact process of inducing structure from a musical sequence in an algorithmic fashion, e.g. Schenkarian analysis \cite{neumeyer_guide_2018}. If we wish to leverage the power afforded by state-of-the-art procedural generation algorithms in our musical exercise generator, we must utilize a second induction process $I_2$ that is capable of partitioning a musical sequence into meaningful, labeled segments for use as examples in the dataset $\pazocal{D}$. Without $I_2$, the parent induction process $I$, which generates the set of rules $R$, would not function correctly as it requires examples $d_i \in \pazocal{D}$ to have labeled structure. It is important to note that $I$ and $I_2$ generally serve different purposes within the context of a procedural generation algorithm: $I$ finds a set of rules $R$ that generalize over multiple sequences; $I_2$ emits a ``structural annotation'' that categorizes each element in a single sequence. 
 
Here we evaluate five grammar induction algorithms as candidates for implementing the induction process $I_2$: \textsc{LZ78} \cite{ziv_universal_1978}, \textsc{Repair} \cite{larsson_offline_1999}, \textsc{LongestFirst} \cite{wolff_algorithm_2011}, \textsc{MostCompressive} \cite{nevill-manning_-line_2000}, and  \textsc{Sequitur} \cite{nevill-manning_identifying_1997}.  One important aspect to consider is how musical sequences should be represented to these algorithms since there is no standard way to encode a musical sequence into symbolic form. We analyze several combinations of musical viewpoints \cite{conklin_feature_2008} and their effect on the behavior of each algorithm. We evaluate the performance of these algorithms and viewpoint combinations against three separate datasets which contain musical sequences with manually annotated phrase segments.\footnote{All code and data can be found at \url{https://github.com/reedperkins/grammatical-induction-phrase-segmentation}
}

\section{Methodology}\label{chp:chapter3}
Our goal is to evaluate grammatical induction algorithms on the task of musical phrase segmentation. Here we provide a summary of the datasets used in our evaluation and the form of the musical sequences that will be used as input for the algorithms as well as ground truth annotations that we use to quantitatively evaluate an algorithm's performance.

\subsection{Data}\label{sec:Data}
The musical sequences and annotations for our experiments are sourced from three different datasets: the Johannes Kepler University Patterns Development Dataset, the Essen Folksong Collection, and the Latter-day Saint Hymns Annotation Collection. Each dataset differs in the average length of their musical sequences, the genre of music they contain, and how their annotations were collected.

The Johannes Kepler University Patterns Development Dataset (JKUPDD) is the product of the Department of Computational Perception at Johannes Kepler University.\footnote{See \url{https://www.tomcollinsresearch.net/mirex-pattern-discovery-task.html} for an overview of the JKUPDD} It consists of five classical piano pieces --- one each from the composers Mozart, Bach, Chopin, Gibbons, and Beethovan. Labeled phrase annotations for each piece are sourced from expert musical analysis found in \cite{barlow_dictionary_1953,schoenberg_fundamentals_1967,bruhn_js_1993}. The JKUPDD contains representations of each piece in the visual, audio, and symbolic domains. In addition, there are polyphonic and monophonic versions of the audio and symbolic representations. We use only the symbolic monophonic sequences and encode them into our intermediate representation (Sec.~\ref{corr}).

The Essen Folksong Collection (Essen), in its current form, is a collection of around 10,000 monophonic melodies from several different areas of the world \cite{schaffrath_essen_1995}. The Essen dataset was originally encoded in a custom format\footnote{\url{http://www.esac-data.org/}}, although we use a version of the dataset that has been encoded into ABC notation which allows for easier parsing.\footnote{\url{https://ifdo.ca/~seymour/runabc/esac/esacdatabase.html}} The standard convention in the literature maintains that visual line breaks of an encoded tune in the Essen dataset represent unlabeled phrase boundaries \cite{bod_probabilistic_2001,brinkman_exploring_2020}, so we employ the following steps to obtain labeled phrase segment annotations from the raw data:
\begin{enumerate}
    \item Merge tunes from each subcategory file into a single file.
    \item For each tune, apply the following steps:
    \begin{enumerate}
        \item Split the tune based on visual line breaks. Each line is considered a phrase occurrence.
        \item Group equivalent phrase occurrences together. Phrase equality is determined by comparing the midi pitch value and duration of each note.
        \item Assign each phrase group a unique label.
    \end{enumerate}
\end{enumerate}

This procedure yields 3,718 tunes ($\sim 36\%$ of the total dataset size) that contain at least one repeated pattern. We use this subset of tunes in our experiments.

Finally, we include the Latter-day Saint Hymns Annotation Collection (Hymn dataset).\footnote{The Hymns dataset can be found at \url{https://github.com/reedperkins/hymns-dataset}}  The Hymns dataset was constructed by randomly sampling 20 hymns that conformed to standard SATB voicing from the 1985 edition of The Hymnbook of the Church of Jesus Christ of Latter-day Saints \shortcite{the_church_of_jesus_christ_of_latter-day_saints_hymns_1985}. To create a repository of monophonic sequences, the soprano voice line (consisting of midi pitch values and duration values) was extracted from each hymn. These sequences were then presented to university music students as annotation assignments where each student was tasked with formally annotating the phrase segments of 2 different sequences. Each of the 20 musical sequences received 5 phrase segment annotations from different annotators.

\subsection{Sequences and Annotations}\label{corr}

We extract from each dataset a series of $(\omega^\pazocal{E}_i, P_i)$ pairs, where each $\omega^\pazocal{E}_i$ is a monophonic musical sequence and $P_i$ is a corresponding ground truth annotation created by human annotators. The raw musical sequences contained in each dataset are encoded in different formats and the total amount of musical information conveyed by a given sequence is limited as a result of its encoding format and other factors. For example, tunes in the Essen dataset have sporadic metadata with respect to key signature and time signature information that is, in some cases, incorrect or missing completely. The raw XML for the Hymns dataset contains correct time signature information and note ornamentations (such as fermatas, slurs, etc.) but also contains errors in the key signature data across all hymns. To reconcile the differences between these formats and their representational limits, we convert the musical sequences in each dataset into sequences of 3-tuples $(o, p, d)$, which we will refer to as \textit{note events}, where $o$, $p$ and $d$ represent the onset time, the midi pitch value, and  duration (in quarter notes). Extra information such as ornamentation and key and time signature is not included. As an example, Figure~\ref{fig:noteevents} in the Appendix depicts the note event sequence $\omega^\pazocal{E} \in \pazocal{E}^*$ that contains each note event $e_1, e_2, ..., e_n \in \pazocal{E}$ for Hymn 5 from the Hymns dataset.

In addition to extracting musical sequences from each dataset, we also extract a ground truth annotation $P$ for each sequence. An annotation $P$ is a set of phrases, each of which is a  contiguous subsequence of note events that are grouped under the same semantic label (supplied by the annotator). More formally, an individual phrase $p \in P$ is defined as a set $O$ of occurrences, where each occurrence $o \in O$ is equivalent to the range $[x_1,\ x_2]$ such that $x_1 < x_2 \in \mathbb{N}$. As an example, Figure~\ref{fig:ann1example} in the Appendix shows a ground truth phrase annotation for Hymn 2. A dataset $\pazocal{D}$ is a list of 2-tuples $d_i = \left( \omega^\pazocal{E}_i, P_i \right)$ where $\omega^\pazocal{E}_i \in \pazocal{E}^*$ is a sequence of note events with the corresponding ground truth annotation $P_i$. We use the terms ``phrases'' and ``annotation(s)'' interchangeably.

\section{Algorithms}\label{sec:algs}
The grammatical induction algorithms we use fall into two classes: online algorithms that process a sequence token by token and emit a grammar at each step, and offline algorithms that perform a transformation on the entire sequence according to some global optimization function. Traditionally, each of these algorithms were designed to consume sequences of Unicode characters for data compression applications. We regard the actual contents of an input sequence to be an implementation detail, and generalize these algorithms to work on sequences of arbitrary elements, including feature vectors created from musical viewpoints (see Section~\ref{sec:vps} for details).

\subsection{LZ78}\label{ss:lz}

\begin{algorithm}[t]
\caption{LZ78 Algorithm for Generic Sequences}
\label{alg:lzw}
\begin{algorithmic}[1]
\Require{A sequence $\omega$} 
\Procedure{LZ78}{$\omega$}
    \State $G \gets \{ N_s \rightarrow ()\}$
    \State $\beta \gets ()$
    \For{$t$ in $\omega$}
        \State $\beta \gets \beta \bigoplus t$
        \If{$\exists r \in G \text{ s.t. } \fn{rhs}(r) = \beta$}
            \State$\beta \gets \fn{lhs}(r)$
        \Else
            \State $G \gets G/\{N_s \rightarrow \alpha\} \cup \{N_s \rightarrow \alpha \bigoplus N_\beta\} \cup \{ N_\beta \rightarrow \beta \}$
            \State $\beta \gets ()$
        \EndIf
    \EndFor
    \State $G \gets G/\{N_s \rightarrow \alpha\} \cup \{N_s \rightarrow \alpha \bigoplus \beta\}$
    \State\Return $G$
\EndProcedure
\end{algorithmic}
\end{algorithm}

The \lz{} algorithm \cite{ziv_universal_1978} is an online compression algorithm that utilizes a special lookup table to substitute previously seen phrases in the input with a shorter representation. We present a slightly altered version of the algorithm that operates on a grammar directly, making it a proper grammatical induction algorithm, in Algorithm~\ref{alg:lzw}. It processes the input in a single pass as follows. First, a rule is created with $N_s$ as the left-hand side and an empty sequence as the right-hand side and added to an empty grammar $G$ (Line 2). Next, while there is still input left to process, the current token $t$ is appended to a buffer $\beta$ (Line 5). If the contents of the buffer match the right-hand side of any rule in $G$, the buffer is replaced with the left-hand side of that rule and the algorithm continues (Lines 6--7). Otherwise, a new non-terminal symbol $N_\beta$ is appended to the right-hand side of the start rule $N_s$ and a new rule using $N_\beta$ as the left-hand side and $\beta$ as the right-hand side is added to $G$ (Line 9) and the buffer is cleared (Line 10). After the input sequence is completely consumed, the contents of the buffer (which may be empty) is appended to right-hand side of the start rule (Line 11). This formulation of \lz{} makes use of two additional functions: $\fn{lhs} :\pazocal{R} \rightarrow \pazocal{N}$ returns the left-hand side of a rule; $\fn{rhs} : \pazocal{R} \rightarrow \{ \pazocal{T} \cup \pazocal{N} \}^*$ returns the right-hand side of a rule.

\subsection{Sequitur}\label{ss:seq}

\sequitur{} \cite{nevill-manning_identifying_1997} is another algorithm in the family of online compression algorithms that iteratively builds a context-free grammar as it consumes an input sequence $\omega \in \pazocal{T}^*$. Unlike \lz{}, \sequitur{} employs a set of transformations that ensure two constraints are met: 
\begin{enumerate}
    \item No pair of symbols occurs more than once in the right-hand side of any rule (\textit{digram uniqueness}).
    \item Every non-terminal that appears in the right-hand side of each rule appears at least twice (\textit{rule utility}).
\end{enumerate}
These constraints ensure that every non-terminal captures a repeated pattern in the original input sequence, and that any repeated pattern will be represented by a rule in the grammar. 

\begin{algorithm}[t]
\caption{Sequitur Algorithm for Generic Sequences}
\label{alg:sequitur}
\begin{algorithmic}[1]
\Require{A sequence $\omega$} 
\Procedure{Sequitur}{$\omega$}
    \State $G \gets \{N_s \rightarrow ()\}$
    \For{$t$ in $\omega$}
        \State $G \gets G/\{N_s \rightarrow \alpha\} \cup \{N_s \rightarrow \alpha \bigoplus t\}$
        
        \State $done \gets False$
        \While{not $done$}
            \State $done \gets True$
            
            \State $dgrams \gets \{ d \mid d \in \fn{repeats}(G) \wedge \fn{len}(d) = 2 \}$
            \For{$d$ in $dgrams$} 
                \State $done \gets False$
                \State $G \gets G_{d \Rightarrow N_d} \cup  \{ N_d \rightarrow d \} $
                \State $dgrams \gets \{ d \mid d \in \fn{repeats}(G) \wedge \fn{len}(d) = 2\}$
            \EndFor
            
            \State $unused \gets \{ r \mid r \in G \wedge \fn{count}(\fn{lhs}(r), G) < 2 \}$
            \For{$\rho$ in $unused$} 
                \State $done \gets False$
                \State $G \gets G_{\fn{lhs}(\rho) \Rightarrow \fn{rhs}(\rho)}$
                \State $G \gets G/\{\rho\}$
                \State $unused \gets \{ r \mid r \in G \wedge \fn{count}(\fn{lhs}(r), G) < 2 \}$
            \EndFor
        \EndWhile
    \EndFor
    \State \Return $G$
\EndProcedure
\end{algorithmic}
\end{algorithm}

Algorithm~\ref{alg:sequitur} gives pseudocode for \sequitur{} that makes use of several auxiliary functions. The functions \fn{lhs} and \fn{rhs} are defined in Section~\ref{ss:lz}. We define a function $\fn{len} : \{ \pazocal{T} \cup \pazocal{N} \}^* \rightarrow \mathbb{N}$ that simply returns the length of a sequence of symbols $\omega \in \{ \pazocal{T} \cup \pazocal{N} \}^*$. The function $\fn{repeats} : \pazocal{G} \rightarrow \mathcal{P}(\{ \pazocal{T} \cup \pazocal{N} \}^*)$ maps a grammar $G \in \pazocal{G}$ to the set of non-overlapping subsequences that occur more than once in the right-hand sides of $G$. The function $\fn{count} : \{ \pazocal{T} \cup \pazocal{N} \}^* \times \pazocal{G} \rightarrow \mathbb{N}$ returns the number of non-overlapping occurrences of a sequence $\omega \in \{ \pazocal{T} \cup \pazocal{N} \}^*$ in the right-hand sides of a grammar $G \in \pazocal{G}$.  We define a substitution operator $G_{x \Rightarrow y}$ which replaces all $x$s with $y$s in right-hand sides of $G$.

\sequitur{} works as follows.\footnote{The original formulation of \sequitur{} is more complex and guarantees linear time complexity. Because we are dealing with sequences that are only hundreds of elements long (\sequitur{} was originally designed to handle millions), we provide an alternative description of \sequitur{} that sacrifices worst-case asymptotic performance for a more straightforward implementation.} First, a rule is created with $N_s$ as the left-hand side and an empty sequence as the right-hand side and added to an empty grammar $G$ (Line 2). Then, while there is still input to process, the following actions are performed. The current token $t$ is appended to the end of the right-hand side of the start rule (Line 4). To ensure that $G$ meets the first constraint of digram uniqueness, the set of non-overlapping repeated phrases of length 2 are computed and stored in a variable $digrams$ (Line 8). Beginning with the first of these digrams, $d$, a new non-terminal symbol $N_d$ is created, after which all occurrences of $d$ are substituted with $N_d$ (denoted as $G_{d \Rightarrow N_d}$), and, finally, a new rule using $N_d$ as the left-hand side and $d$ as the right-hand side is added to $G$ (Line 11). The substitution process will change the number of occurrences of some digrams, and thus $digrams$ will need to be recomputed (Line 12). This process is repeated until there are no digrams in the right-hand sides of $G$ that occur more than once, which satisfies the first constraint.

Next, to ensure that $G$ meets the second constraint of digram utility, the set of rules in $G$ whose left-hand side non-terminal symbol occurs at most once in the right-hand sides of $G$ are computed and stored in a variable $unused$ (Line 13). These rules are under-utilized. The first of these rules, $\rho$, is removed by first substituting all occurrences (that is, the single occurrence) of $\rho$'s left-hand side non-terminal symbol with its right-hand side (Line 16). Afterwards, $\rho$ is removed from $G$ (Line 17). This substitution will also change the number of occurrences of some digrams, and therefore $unused$ will also need to be recomputed (Line 18). Additionally, the back-substitution that occurs in Line 16 could cause some digrams to occur more than once, which would place $G$ in violation of the first constraint. After the second constraint is met, Lines 9-18 will need to be repeated until no changes to $G$ are made. After this point, $G$ has met both constraints and the next token in the input sequence can be processed. When the input sequence is exhausted, $G$ is returned (Line 19). 

\subsection{Repair, Most Compressive, Longest First}
\begin{algorithm}[t]
\caption{IRR Algorithm for Generic Sequences}
\label{alg:irr}
\begin{algorithmic}[1]
\Require{A sequence $\omega$ and a score function $f$}
\Procedure{IRR}{$\omega$, $f$}
    \State $G \gets \{N_s \rightarrow \omega\}$
    \While{$\exists \omega' \text{ s.t. } \omega' \gets \argmax_{\omega' \in \fn{repeats}(G)} f(\omega', G)$\newline
        \hspace*{4em} $\wedge \text { } |G_{\omega' \Rightarrow N_{\omega'}}| < |G|$}
        \State $G \gets G_{\omega' \Rightarrow N_{\omega'}} \cup \{ N_{\omega'} \rightarrow \omega' \}$
    \EndWhile
    \State \Return $G$
\EndProcedure
\end{algorithmic}
\end{algorithm}

In contrast to the online algorithms introduced above, offline algorithms work by operating on the entire input sequence and substituting subsequences with non-terminal symbols until no new productions can be made. Carrasscosa grouped three offline grammatical induction algorithms into a unified framework known as the Iterative Repeat Replace (\textsc{IRR}) algorithm \cite{dediu_choosing_2010}. We present our version of the algorithm in Algorithm~\ref{alg:irr}. We implement all three versions of \textsc{IRR} where each instance uses a different objective function. The first objective function uses the auxiliary functions \fn{repeats} and \fn{count} (defined in Section~\ref{ss:seq}) and yields the subsequence $\omega' \in \{ \pazocal{T} \cup \pazocal{N} \}^*$ that occurs most frequently in the right-hand sides of $G$: 

\begin{equation} \label{eqn:repair}
\omega' = \argmax_{\omega' \in \fn{repeats}(G)} \fn{count}(\omega', G)
\end{equation}

\noindent \textsc{IRR} instantiated with Equation~\eqref{eqn:repair} is also known as \repair{}  \cite{larsson_offline_1999}. The second objective function returns the longest non-overlapping repeated subsequence in the right-hand sides of $G$ and is known as \longestFirst{}  \cite{wolff_algorithm_2011}:

\begin{equation} \label{eqn:longestFirst}
\omega' = \argmax_{\omega' \in \fn{repeats}(G)} |\omega'|
\end{equation}

\noindent The third objective function returns the non-overlapping repeated subsequence that, when replaced with a non-terminal, would result in the largest reduction in the size of $G$:

\begin{multline} \label{eqn:mostCompressive}
\omega' = \argmax_{\omega' \in \fn{repeats}(G)} |\omega'| \cdot \fn{count}(\omega', G)\\ - |\omega'| - \fn{count}(\omega', G) - 1
\end{multline}

\noindent Equation~\ref{eqn:mostCompressive}, which is the objective function used in \mostCompressive{}  \cite{nevill-manning_-line_2000}, calculates the number of symbols that will be replaced by a non-terminal, plus the cost of adding a non-terminal in each ``hole'' and creating a new rule in the grammar. 

\textsc{IRR} will continue to run while the objective function returns a valid subsequence $\omega'$ and the total size of the grammar after substituting all occurrences of $\omega'$ with a new non-terminal symbol in the grammar, denoted by $|G_{\omega' \Rightarrow N_{\omega'}}|$, is smaller than the size of current grammar, denoted by $|G|$. We follow \cite{dediu_choosing_2010} in defining the size of a grammar $G$ to be equivalent to the length of the sequence formed by joining the right-hand sides of each rule in the grammar with a special delimiter symbol.

\section{Viewpoints}\label{sec:vps}

Viewpoints are functions that convert a note event $e \in \pazocal{E}$ (defined in Section~\ref{corr}) into a singular value \cite{conklin_feature_2008}. For example, the pitch viewpoint, denoted by the function $\fn{pitch} : \pazocal{E} \rightarrow \{0,1,\dots,127\}$, returns the midi pitch value for a note event $e \in \pazocal{E}$. The duration viewpoint function $\fn{duration}:\pazocal{E} \rightarrow \mathbb{R}^+$ computes the duration (in quarter notes) of a note event $e \in \pazocal{E}$. The \fn{pitch} and \fn{duration} viewpoint functions are both unary functions, and we define additional binary viewpoints that compute inter-onset interval (\fn{ioi}), pitch contour (\fn{pitchC}), and pitch interval (\fn{pitchI}) between two note events as follows:

\begin{align}
\fn{ioi}(e_{i-1}, e_{i}) &=
  \begin{cases}\label{eqn:ioi}
    \bot \quad \text{ if $e_{i-1}$ is $\bot$} \\
    \fn{duration}(e_{i}) - \fn{duration}(e_{i-1})\\
        \quad\quad \text{otherwise}
\end{cases}\\
\fn{pitchC}(e_{i-1}, e_{i}) &=
  \begin{cases}\label{eqn:pitchC}
      \bot & \text{if $e_{i-1}$ is $\bot$} \\
      1 & \text{if $\fn{pitch}(e_{i}) > \fn{pitch}(e_{i-1})$} \\
      0 & \text{if $\fn{pitch}(e_{i}) = \fn{pitch}(e_{i-1})$} \\
      -1 & \text{if $\fn{pitch}(e_{i}) < \fn{pitch}(e_{i-1})$} \\
  \end{cases}\\
\fn{pitchI}(e_{i-1}, e_{i}) &=
  \begin{cases}\label{eqn:pitchI}
      \bot \quad\ \text{ if $e_{i-1}$ is $\bot$} \\
      \fn{pitch}(e_{i}) - \fn{pitch}(e_{i-1})\\
        \quad\quad \text{otherwise} \\
  \end{cases}
\end{align}

\noindent where $e_i$ and $e_{i-1}$ represent note events at sequence indices $i$ and $i-1$, respectively, and $\bot$ represents the null value.  We define a function $\fn{transform} : \pazocal{V} \times \pazocal{E}^* \rightarrow \Phi^*$ that converts an $n$\nobreakdash-length sequence of note events $\omega^\pazocal{E} \in \pazocal{E}^*$ into a sequence of feature vectors $\omega^\Phi \in \Phi^*$  by applying a \textit{viewpoint combination} (i.e. a set $V \in \pazocal{V}$ of multiple viewpoint functions) to each note event in the sequence. More specifically, each note event $e_1, e_2, ..., e_n \in \omega^\pazocal{E}$ is mapped to its corresponding feature vector $\phi_1, \phi_2, ..., \phi_n \in \omega^\Phi$, where each $\phi_i \in \{\bot \cup \mathbb{N} \}^{|V|}$ is formed by applying the constituent viewpoint functions of $V \in \pazocal{V}$ to $e_i$ (e.g. $v_1(e_i)$, $v_2(e_i)$, \dots, $v_{|V|}(e_i)$) and concatenating the results into a $|V|$\nobreakdash-dimensional feature vector.  In our experiments, $\pazocal{V} = \mathcal{P}($\texttt{\{ pitch, duration, ioi, pitchI, pitchC \}}$)$ where $\mathcal{P}$ denotes the powerset. Figure~\ref{fig:vpexample} in the Appendix shows the  result of \fn{transform} supplied with the note event sequence of Hymn 5 from the Hymns dataset and Viewpoint Combination Index (VCI) 31. Each of the algorithms described in this section consumes a sequence of feature vectors $\omega^\Phi \in \Phi^*$ produced by the \fn{transform} function as input.

\section{Experimental Setup}\label{sec:Experiment}

Our goal is to see which algorithms and viewpoint combinations produce the best labeled segments for different genres of musical sequences.  We run $|\pazocal{D} \times \pazocal{V} \times \pazocal{A}|$ simulations, where $\pazocal{D}$ is the union of all three datasets, $\pazocal{V}$ is the set of viewpoint combinations, and $\pazocal{A}$ is the set of algorithms.\footnote{The JKUPDD and the Hymns dataset contain 5 and 20 musical sequences, respectively. Due to computational resource constraints, we only used 1,000 sequences from the Essen dataset, which we sampled from a uniform random distribution. In total this amounted to $(1000+5+20) * 31 * 5 = 158,875$ simulations.} Recalling our definition from Section~\ref{corr}, a dataset $\pazocal{D}$ comprises 2-tuples $d_i = (\omega^\pazocal{E}_i, P_i)$ that each contain a note event sequence $\omega^\pazocal{E}_i$ and a corresponding set $P_i$ of ground truth patterns. Each simulation involves running an algorithm $a \in \pazocal{A}$ over a viewpoint-transformed sequence $\omega^{\Phi}_i = \fn{transform}(\omega^\pazocal{E}_i, V)$ for some $V \in \pazocal{V}$ and then extracting a set of discovered patterns $Q$ from the resulting grammar $G = a(\omega^{\Phi}_i)$. We extract $Q$ from $G$ by recursively expanding all non-terminals in the right-hand side of the $G$'s start rule---each distinct non-terminal $N_j$ expands into a subsequence $\omega^{\Phi}_j$, which we interpret as a musical phrase with label $j$. Occurrences of phrase $j$ are discovered by finding the starting and ending indices of each subsequence $\omega^\Phi_j$.

\begin{algorithm}[t]
\caption{Fuzzy Intersection}
\label{alg:fuzzyIntersection}
\textbf{Require:} Sets $P$ and $Q$ of patterns, a sequence of feature vectors $\omega^\Phi$, and a threshold value $\tau$. \\
\textbf{Output:} The tuple $(|A|, |B|)$ where $|A|$ and $|B|$ are the number of elements in $P$ that matched elements in $Q$ with a score greater than or equal to $\tau$, and vice versa.
\begin{algorithmic}[1]
\Procedure{FuzzyIntersection}{$P$, $Q$, $\omega^\Phi$, $\tau$}
    \State $A \gets \{\}$
    \State $B \gets \{\}$
    \For{$(p, q) \in P \times Q$}
        \For{$(o_p, o_q) \in p \times q$}
            \State $\omega_p \gets \fn{extract}(o_p, \omega^\Phi)$
            \State $\omega_q \gets \fn{extract}(o_q, \omega^\Phi)$
                \If{$\fn{similarity}(\omega_p, \omega_q) \geq \tau$}
                \State $A \gets A \cup \{p\}$
                \State $B \gets B \cup \{q\}$
        \EndIf
        \EndFor
    \EndFor
    \State \Return $|A|, |B|$
\EndProcedure
\end{algorithmic}
\end{algorithm}

Pattern similarity between a ground truth annotation $P$ and a discovered annotation $Q$ is measured by $|P \cap Q|$. Computing the intersection of $P$ and $Q$ based on strict equality, however, would unfairly penalize algorithms that produce segmentations that are very close to the ground truth but do not match it exactly. Instead, we  define the \textsc{FuzzyIntersection} algorithm (Algorithm~\ref{alg:fuzzyIntersection}) that computes $P \cap Q$ by matching patterns $p \in P$ with $q \in Q$ based on a similarity measure.  The similarity between patterns $P$ and $Q$ is determined by computing the maximum similarity score over all combinations of the occurrences of $p \in P$ and $q \in Q$.  We can evaluate the similarity of two occurrences $o_p$ and $o_q$ by first extracting from $\omega^\Phi_i$ the subsequences they demarcate, which we will refer to as $\omega_p$ and $\omega_q$, and then comparing $\omega_p$ and $\omega_q$ using generic sequence comparision algorithms. The extraction process is accomplished using the function $\fn{extract} : \mathbb{N}^2 \times \Phi^* \rightarrow \Phi^*$ which returns the  subsequence of a feature vector sequence $\omega^\Phi \in \Phi^*$ spanned by a pair of integer indices. We implement \fn{similarity} using the Levenshtein edit distance algorithm:

\begin{equation} \label{eqn:levsim}
\fn{similarity}(\omega_p, \omega_q) = 1 - \frac{\fn{Lev}(\omega_p, \omega_q)}{\max (|\omega_p|, |\omega_q|)}
\end{equation}

\noindent where \texttt{Lev} computes the Levenshtein edit distance between two generic sequences---in this case, sequences of feature vectors produced by viewpoint combinations. We use the unit cost of 1 for all edit distance operations.  In summary, we compute $P \cap Q$ by running \textsc{FuzzyIntersection} with arguments $P$, $Q$, $\omega^\Phi_i$ and $\tau = 0.7$.\footnote{For notational simplicity, we assume that \textsc{FuzzyIntersection} maintains global references to \fn{similarity},  and \fn{extract}.}

With Algorithm~\ref{alg:fuzzyIntersection} in place, we are able to compute precision and recall scores for the set of patterns $Q$ discovered by an algorithm against a set of ground truth patterns $P$:

\begin{equation} \label{eqn:prec}
    \fn{precision}(P, Q) = \frac{|P \cap Q|}{|Q|}
\end{equation}

\begin{equation} \label{eqn:rec}
    \fn{recall}(P, Q) = \frac{|P \cap Q|}{|P|}
\end{equation}

\noindent We unify the precision and recall scores of an algorithm into a single number by computing their harmonic mean, also known as the F1 score:

\begin{align} \label{eqn:F1}
    \fn{F1}(P, Q) & = \frac{2}{\frac{1}{\fn{precision}(P, Q)} + \frac{1}{\fn{recall}(P, Q)}}\\ & = 2 \cdot \frac{\fn{precision}(P, Q) \cdot \fn{recall}(P, Q)}{\fn{precision}(P, Q) + \fn{recall}(P, Q)}
\end{align}

\section{Results}\label{chp:results}
Here we present quantitative and qualitative results for  the experiments described in Section~\ref{sec:Experiment}.

\subsection{Experiment Results}

On the Essen dataset the ranking of each algorithm by maximum F1 score over all viewpoint combinations is as follows: \longestFirst{} (maximum F1 score 0.35), followed by \mostCompressive, \sequitur, \repair, and \lz. This ordering is the same for the JKUPDD and the Hymns dataset as well and also applies in each case when considering the average F1 score. When given the correct viewpoint combination, \longestFirst{} performs the best on each dataset. 

The choice of viewpoint combination has a substantial impact on an algorithm’s F1 score. On the Hymns dataset, VCI\nobreakdash-2 (\{ \fn{duration} \}) and VCI\nobreakdash-3 (\{ \fn{ioi} \}) produced the highest F1 scores for each algorithm. On the Essen dataset, VCI\nobreakdash-2 produced the maximum F1 score for each algorithm. On the JKUPDD, the story is much more varied. The viewpoint that generated the maximum F1 score was VCI\nobreakdash-5 (\{ \fn{pitchI} \}) for \lz{}, VCI\nobreakdash-2 for \repair{}, VCI\nobreakdash-18 (\{ \fn{duration}, \fn{pitch}, \fn{pitchI} \}) for \mostCompressive{} and \sequitur{}, and VCI\nobreakdash-12 (\{ \fn{duration}, \fn{pitchI} \}) for \longestFirst{}. We conclude from these results that the \fn{duration} viewpoint, either by itself or combined with other viewpoints, is the most important viewpoint to consider when using the grammatical induction algorithms. 

The fact that \fn{pitchI} was a component viewpoint for each of the highest scoring viewpoint combinations per algorithm on the JKUPDD suggests that the JKUPDD contains pattern occurrences that are transpositions of a canonical pattern. The \fn{pitch} viewpoint precludes an algorithm from capturing transposed patterns, but \fn{pitchI} is invariant to transposition.

The content of the datasets themselves affects the F1 score of each algorithm. All algorithms received the highest maximum and average F1 score on the JKUPDD. This is likely due to the fact that the JKUPDD contains patterns with more (and longer) occurrences than the Essen and Hymns datasets. This phenomenon can be observed quantitatively by studying the variance of F1 scores over the viewpoint combinations for a given algorithm on the dataset. \longestFirst{} on the JKUPDD has the highest average F1 score ($\mu = 0.58$) and the lowest variance ($\sigma = 0.04)$. In contrast, the same algorithm on the Hymns dataset has the highest variance ($\sigma = 0.12$). On the JKUPDD, almost every VCI has a good chance of scoring highly, whereas on the Hymns and Essen datasets the choice of viewpoint is much more important.  See Figure~\ref{fig:f1Results} in the Appendix for additional detail.

\subsection{Qualitative Analysis}

For the Hymns dataset, we can also qualitatively compare the output of the five algorithms with the annotations produced by the five human annotators, revealing not only information about algorithm performance but also the degree to which the human annotators were in agreement.  As an example, we arbitrarily chose Hymn 2 to analyze.  (Figure~\ref{fig:hymn2} in the Appendix shows ten annotations for Hymn 2: five from human annotators and five from the grammatical induction algorithms.)

The ground truth annotations for Hymn 2 reveal interesting discrepancies between annotators. Each annotator produced phrase labels that covered the entire sequence. However, Annotators 2--5 chose to segment the sequence into four 4-bar segments whereas Annotator~1 divided the sequence into eight 2-bar segments. These divisions carry important implications. Annotator~1, for example, focused on identifying smaller patterns --- they identified that bars 3--4 and 7--8 are exact repetitions of one another, and subsequently gave these segments the same label. Additionally, they identified bars 13--14 as an embellishment of the first two opening bars. In total, Annotator~1 identified eight phrases as well as two patterns. The rest of the annotators, on the other hand, were more concerned with phrase identification on a broader scale. In contrast with Annotator 1, Annotators 2--5 identified four 4-bar phrases that spanned bars 1--4, bars 5--8, bars 9--12, and bars 13--16. There was a high level of consensus among these four annotators; Annotators 2 and 5 agreed that bars 1--4 and 4--8 were repeated phrases and Annotators 3 and 4 produced the exact same annotation. The bifurcation of the annotators on this particular sequence becomes more apparent when we compare the F1 scores between their annotations as determined using Algorithm~\ref{alg:fuzzyIntersection} with $\tau = 0.7$.  The annotations provided by Annotator 1 failed to match any of those given by Annotators 2--5 and all the phrase boundaries produced by Annotators 2--5 were determined to match (though the labeling differs slightly).  See Figure~\ref{fig:f1Comp} in the Appendix for details.

The annotations produced by the grammatical induction algorithms both align with and diverge from the human annotators (and each other) in interesting ways. It is immediately clear that the phrases identified by \lz{} do not appear to be musically significant compared to those found by the other annotators and algorithms. In concordance with Annotator~1, on the other hand, \repair{} identified bars 3--4 and 7--8 as the first two instances of Pattern 2, which match the phrases included in Pattern B (for Annotator 1). However, \repair{} also included bars 11--12 in Pattern 2, whereas Annotator 1 labeled bars 11--12 as the singleton Phrase E. It is likely that Annotator 1 was interpreting Phrase E in the context of the underlying chord progression, whereas \repair{} in this case is only seeing the duration values of each note event. Using the duration viewpoint allowed \repair{} to capture one of the patterns identified by Annotator 1, at the cost of possibly over-estimating the number of occurrences by one additional phrase. Furthermore, the phrases produced by \repair{} exhibit some interesting characteristics; each of Annotator 1's phrase boundaries are either captured directly (e.g. Phrase B $\approx$ Phrase 2) or by composition (e.g. Phrase F $\approx$ Phrase 4 + Phrase 3) in \repair{}'s annotation. Examining the pairwise F1 scores between all the annotations for Hymn 2 shows that, despite this, the match score between \repair{} and Annotator 1 (0.31) was lower than that of \longestFirst{} and  Annotator 1 (0.57). In this case, \repair{} is not able to score as high as \longestFirst{} because most of the phrases it found are by themselves too small to match the phrases from Annotator~1.  See Figure~\ref{fig:f1Comp} in the Appendix for detail.

On the other hand, \sequitur{}, \mostCompressive{}, and \longestFirst{} matched much more closely with Annotators 2--5, at least in the first eight bars. Each of these algorithms divided the first eight bars into two 4-bar segments and assigned the same phrase label to each. This allowed them to match the  phrases in bars 1--4 and 5--8 discovered by Annotators 2--5.  In bars 9--16 however, these algorithms failed to capture any of the longer 4-bar phrases that were identified by Annotators 2--5. This phenomenon demonstrates one of the tradeoffs of using these particular grammatical induction algorithms for musical phrase segmentation: these algorithms can efficiently detect patterns and encode them in a human-readable way, but each occurrence of a repeated pattern must contain the exact same number of notes. The use of musical viewpoints is intended to mitigate the rigidity of this constraint to some degree with respect to transposition, but one hard limitation is that melodic embellishment, either by expansion or reduction, will remain undetected.  According to Annotator 5, bars 13--16 can be seen as an embellishment of bars 1--4 and 5--8, but \longestFirst{}, for example, would not be able to include bars 13--16 as an occurrence of Pattern 1 because that phrase contains more notes than the first two occurrences of Pattern 1. This is why \sequitur{}, \mostCompressive{}, and \longestFirst{} are unable to find the longer phrases in bars 9--16 identified by Annotators 2--5 (although \sequitur{} did manage to identify Phrase 4 which matched Annotator 1's Phrase E). 

\begin{figure}[t]
    \centering
    \begin{subfigure}[]{\linewidth}
        \centering
        \begin{tabular}{cl}
            $N_s \rightarrow$ & $N_1,N_1,N_2,N_3,N_4,N_5,N_2,\langle1.0\rangle,N_6,N_6,N_3$\\
            $N_1 \rightarrow$ & $N_2,N_2,N_4$\\
            $N_2 \rightarrow$ & $N_7,N_7$\\
            $N_3 \rightarrow$ & $N_7,\langle2.0\rangle$\\
            $N_4 \rightarrow$ & $N_5,N_9,\langle1.0\rangle,\langle2.0\rangle$\\
            $N_5 \rightarrow$ & $N_8,N_9$\\
            $N_6 \rightarrow$ & $N_8,N_8$\\
            $N_7 \rightarrow$ & $\langle1.0\rangle,\langle1.0\rangle$\\
            $N_8 \rightarrow$ & $\langle0.5\rangle,\langle0.5\rangle$\\
            $N_9 \rightarrow$ & $N_8,\langle1.0\rangle$
        \end{tabular}
    \end{subfigure}
    \caption[The grammar $G$ for Hymn 2 generated by \sequitur{} using the \fn{duration} viewpoint]{The grammar $G$ for Hymn 2 generated by \sequitur{} using the \fn{duration} viewpoint. Numbers encased in angle brackets are the feature vectors $\phi_i$ which each contain a single duration value. These feature vectors are also the terminal symbols for $G$.}
    \label{fig:gram}
\end{figure}

Thus far, we have only analyzed the performance of grammatical induction algorithms with respect to ground truth data. Our method of extracting annotations from the grammars produced by each of the grammatical induction algorithms involves expanding only the non-terminals that are found in the right-hand side of the start rule --- this was done to ensure that the annotations emitted by a grammatical induction algorithm would consist of disjoint phrases. In doing so,  we actually overlook one of the most interesting properties of grammatical induction algorithms: their ability to produce hierarchical structures. Figure~\ref{fig:gram} shows the  grammar $G$ emitted by \sequitur{} on Hymn 2 using the \fn{duration} viewpoint.  Visualizing the full grammar produced by \sequitur{} reveals how the final phrase annotations demarcated by the non-terminals in the right-hand side of $N_s$ are actually composed from smaller phrases or musical motifs (See Figure~\ref{fig:hier} in the Appendix for a visual interpretation of $G$ superimposed on the Hymn 2). Unfortunately, most ground truth data for musical sequence annotations contains disjoint phrase labels, which is why we based the musical phrase segmentation task as well as  our evaluation metrics on the assumption that the annotations we extracted from the grammatical induction algorithms would contain disjoint phrases. That said, a future procedural generation algorithm could leverage the fact that grammatical induction algorithms expose the compositional nature of musical phrases.

\section{Conclusion}\label{chp:conclusion}
Our goal was to evaluate the performance of grammatical induction algorithms on the task of musical sequence segmentation using five different algorithms that infer context-free grammars from musical sequences. We created a set of 31 different viewpoint combinations out of 5 distinct viewpoint functions so that we could measure the degree to which the representation of a musical sequence affects the performance of an algorithm. The sequences for this experiment were sourced from three separate datasets. Finally, we performed an experiment by running all five grammatical induction algorithms over every sequence in the dataset, transforming each sequence by all 31 viewpoint combinations. We measured the performance of a given algorithm and viewpoint-transformed sequence by comparing the patterns output by the grammar against a corresponding set of ground truth patterns.

The results of our experiments show that the best viewpoint combination index was VCI-2, which consists solely of the duration viewpoint.  Notably, this suggests that the annotators themselves were also heavily influenced by the rhythmic patterns in the musical sequences that we studied. If a musical sequence contains patterns which are likely to be transposed, VCI-2 works well when combined with VCI-5. 
\longestFirst{} was the best performing algorithm on all datasets when given sequences transformed by the optimal VCI, attaining a maximum F1 score (averaged over all sequences) of 0.35 with VCI-2 on the Essen dataset, 0.58 with VCI-12 on the JKUPDD, and 0.50 with VCI-3 on the Hymns dataset.


\bibliographystyle{named}
\bibliography{references}

\begin{thebibliography}{}

\bibitem[\protect\citeauthoryear{Barlow and Morgenstern}{1953}]{barlow_dictionary_1953}
Harold Barlow and Sam Morgenstern.
\newblock {\em A {Dictionary} of {Musical} {Themes}}.
\newblock Crown, New York, 1953.

\bibitem[\protect\citeauthoryear{Bod}{2001}]{bod_probabilistic_2001}
Rens Bod.
\newblock Probabilistic {Grammars} for {Music}.
\newblock In {\em Proceedings of the {Belgian}-{Dutch} {Conference} on {Artificial} {Intelligence}}, page~8, 2001.

\bibitem[\protect\citeauthoryear{Bodily and Ventura}{2021}]{bodily_inferring_2021}
Paul Bodily and Dan Ventura.
\newblock Inferring structural constraints in musical sequences via multiple self-alignment.
\newblock In {\em Proceedings of the 43rd Annual Meeting of the Cognitive Science Society}, pages 1112--1118, 2021.

\bibitem[\protect\citeauthoryear{Brinkman}{2020}]{brinkman_exploring_2020}
Andrew Brinkman.
\newblock {\em Exploring the {Structure} of {Germanic} {Folksong}}.
\newblock Doctoral {Dissertation}, Ohio State University, OhioLINK Electronic Theses and Dissertations Center, 2020.

\bibitem[\protect\citeauthoryear{Bruhn}{1993}]{bruhn_js_1993}
Siglind Bruhn.
\newblock {\em J.{S}. {Bach}'s {Well}-{Tempered} {Clavier}: {In}-depth {Analysis} and {Interpretation}}.
\newblock Mainer, Hong Kong, 1993.

\bibitem[\protect\citeauthoryear{Bryden}{2006}]{bryden_using_2006}
K.~Bryden.
\newblock Using a {Human}-in-the-{Loop} {Evolutionary} {Algorithm} to {Create} {Data}-{Driven} {Music}.
\newblock In {\em Proceedings of the 2006 {IEEE} {World} {Congress} on {Computational} {Intelligence}}, pages 2065--2071, Vancouver, CA, 2006.

\bibitem[\protect\citeauthoryear{Carrascosa \bgroup \em et al.\egroup }{2010}]{dediu_choosing_2010}
Rafael Carrascosa, François Coste, Matthias Gallé, and Gabriel Infante-Lopez.
\newblock Choosing {Word} {Occurrences} for the {Smallest} {Grammar} {Problem}.
\newblock In {\em Proceedings of the 4th International Conference on Language and {Automata} {Theory} and {Applications}}, volume LNCS 6031, pages 154--165, 2010.

\bibitem[\protect\citeauthoryear{Cilibrasi \bgroup \em et al.\egroup }{2004}]{cilibrasi_algorithmic_2004}
Rudi Cilibrasi, Paul Vitányi, and Ronald de~Wolf.
\newblock Algorithmic {Clustering} of {Music} {Based} on {String} {Compression}.
\newblock {\em Computer Music Journal}, 28(4):49--67, 2004.

\bibitem[\protect\citeauthoryear{Conklin and Bergeron}{2008}]{conklin_feature_2008}
Darrell Conklin and Mathieu Bergeron.
\newblock Feature {Set} {Patterns} in {Music}.
\newblock {\em Computer Music Journal}, 32(1):60--70, 2008.

\bibitem[\protect\citeauthoryear{Dalhoum \bgroup \em et al.\egroup }{2008}]{dalhoum_computer-generated_2008}
Abdel Latif~Abu Dalhoum, Manuel Alfonseca, Manuel~Cebrián Ramos, Rafael Sánchez-Alfonso, and Alfonso Ortega.
\newblock Computer-{Generated} {Music} {Using} {Grammatical} {Evolution}.
\newblock In {\em Proceedings of the 9th {Middle} {Eastern} {Simulation} {Multiconference}}, pages 55--61, 2008.

\bibitem[\protect\citeauthoryear{Fox}{2006}]{fox_genetic_2006}
Charles Fox.
\newblock Genetic {Hierarchical} {Music} {Structures}.
\newblock In {\em Proceedings of the {Nineteenth} {International} {Florida} {Artificial} {Intelligence} {Research} {Society} {Conference}}, pages 243--247, 2006.

\bibitem[\protect\citeauthoryear{Gilbert and Conklin}{2007}]{gilbert_probabilistic_2007}
Edouard Gilbert and Darrell Conklin.
\newblock A {Probabilistic} {Context}-{Free} {Grammar} for {Melodic} {Reduction}.
\newblock In {\em Proceedings of the {Twentieth} {International} {Joint} {Conference} on {Artificial} {Intelligence}}, pages 83--94, 2007.

\bibitem[\protect\citeauthoryear{Green}{1965}]{green_tonal_1965}
Douglass~M Green.
\newblock {\em Form in Tonal Music: An Introduction to Analysis}.
\newblock Holt, Rinehart, and Winston, New York, 1965.

\bibitem[\protect\citeauthoryear{Guan \bgroup \em et al.\egroup }{2018}]{guan2018melodic}
Yixing Guan, Jinyu Zhao, Yiqin Qiu, Zheng Zhang, and Gus Xia.
\newblock Melodic {Phrase} {Segmentation} by {Deep} {Neural} {Networks}., 2018.
\newblock Available online: https://arxiv.org/abs/1811.05688.

\bibitem[\protect\citeauthoryear{Hernandez-Olivan \bgroup \em et al.\egroup }{2021}]{olivan_boundary_2020}
Carlos Hernandez-Olivan, Jose~R. Beltran, and David Diaz-Guerra.
\newblock {Music Boundary Detection using Convolutional Neural Networks: A Comparative Analysis of Combined Input Features}.
\newblock {\em International Journal of Interactive Multimedia and Artificial Intelligence}, 7(2), 2021.

\bibitem[\protect\citeauthoryear{Kitani and Koike}{2010}]{kitani_improvgenerator:_2010}
Kris~Makoto Kitani and Hideki Koike.
\newblock {ImprovGenerator}: {Online} {Grammatical} {Induction} for {On}-the-{Fly} {Improvisation} {Accompaniment}.
\newblock In {\em Proceedings of the {International} {Conference} on {New} {Interfaces} for {Musical} {Expression}}, pages 469--472, 2010.

\bibitem[\protect\citeauthoryear{Larsson and Moffat}{1999}]{larsson_offline_1999}
N.J. Larsson and A.~Moffat.
\newblock Offline {Dictionary}-based {Compression}.
\newblock In {\em Proceedings of the {Data} {Compression} {Conference}}, pages 296--305, Snowbird, UT, USA, 1999.

\bibitem[\protect\citeauthoryear{Lattner \bgroup \em et al.\egroup }{2015}]{lattner_probabilistic_2015}
Stefan Lattner, Maarten Grachten, Kat Agres, and Carlos~Eduardo Cancino~Chacón.
\newblock Probabilistic {Segmentation} of {Musical} {Sequences} {Using} {Restricted} {Boltzmann} {Machines}.
\newblock In {\em Proceedings of the 5th International Conference on Mathematics and {Computation} in {Music}}, volume LNCS 9110, pages 323--334, 2015.

\bibitem[\protect\citeauthoryear{Louboutin and Meredith}{2016}]{louboutin_using_2016}
Corentin Louboutin and David Meredith.
\newblock Using {General}-purpose {Compression} {Algorithms} for {Music} {Analysis}.
\newblock {\em Journal of New Music Research}, 45(1):1--16, 2016.

\bibitem[\protect\citeauthoryear{Mason \bgroup \em et al.\egroup }{2019}]{mason_lume_2019}
Stacey Mason, Ceri Stagg, and Noah Wardrip-Fruin.
\newblock Lume: {A} {System} for {Procedural} {Story} {Generation}.
\newblock In {\em Proceedings of the 14th {International} {Conference} on the {Foundations} of {Digital} {Games}}, pages 1--9, 2019.

\bibitem[\protect\citeauthoryear{McCormack}{1996}]{mccormack_grammar_1996}
Jon McCormack.
\newblock Grammar {Based} {Music} {Composition}.
\newblock {\em Complex Systems}, 3:321--336, 1996.

\bibitem[\protect\citeauthoryear{Meredith \bgroup \em et al.\egroup }{2002}]{meredith_algorithms_2002}
David Meredith, Kjell Lemström, and Geraint~A. Wiggins.
\newblock Algorithms for {Discovering} {Repeated} {Patterns} in {Multidimensional} {Representations} of {Polyphonic} {Music}.
\newblock {\em Journal of New Music Research}, 31(4):321--345, 2002.

\bibitem[\protect\citeauthoryear{Meredith}{2013}]{meredith_cosiatec_2013}
David Meredith.
\newblock {COSIATEC} and {SIATECCompress}: {Pattern} {Discovery} by {Geometric} {Compression}.
\newblock In {\em Proceedings of the {Music} {Information} {Retrieval} {Evaluation} {eXchange}}, pages 48--54, Curitiba, Brazil, 2013.

\bibitem[\protect\citeauthoryear{Meredith}{2015}]{meredith_music_2015}
David Meredith.
\newblock Music {Analysis} and {Point}-{Set} {Compression}.
\newblock {\em Journal of New Music Research}, 44(3):245--270, 2015.

\bibitem[\protect\citeauthoryear{Müller \bgroup \em et al.\egroup }{2006}]{muller_procedural_2006}
Pascal Müller, Peter Wonka, Simon Haegler, Andreas Ulmer, and Luc Van~Gool.
\newblock Procedural {Modeling} of {Buildings}.
\newblock {\em ACM Transactions on Graphics}, 25(3):614--623, 2006.

\bibitem[\protect\citeauthoryear{Neumeyer}{2018}]{neumeyer_guide_2018}
David Neumeyer.
\newblock {\em A {Guide} to {Schenkerian} {Analysis}}.
\newblock The University of Texas at Austin, 2018.

\bibitem[\protect\citeauthoryear{Nevill-Manning and Witten}{1997}]{nevill-manning_identifying_1997}
C.~G. Nevill-Manning and I.~H. Witten.
\newblock Identifying {Hierarchical} {Structure} in {Sequences}: {A} {Linear}-time {Algorithm}.
\newblock {\em Journal of Artificial Intelligence Research}, 7:67--82, 1997.

\bibitem[\protect\citeauthoryear{Nevill-Manning and Witten}{2000}]{nevill-manning_-line_2000}
C.~G. Nevill-Manning and I.~H. Witten.
\newblock On-line and {Off}-line {Heuristics} for {Inferring} {Hierarchies} of {Repetitions} in {Sequences}.
\newblock {\em Proceedings of the IEEE}, 88(11):1745--1755, 2000.

\bibitem[\protect\citeauthoryear{Nieto and Farbood}{2014}]{nieto_identifying_2014}
Oriol Nieto and Morwaread~M Farbood.
\newblock Identifying {Polyphonic} {Patterns} {From} {Audio} {Recordings} {Using} {Music} {Segmentation} {Techniques}.
\newblock In {\em Proceedings of the 15th {International} {Society} for {Music} {Information} {Retrieval} {Conference}}, pages 411--416, Taipei, Taiwan, 2014.

\bibitem[\protect\citeauthoryear{Ritchie \bgroup \em et al.\egroup }{2015}]{ritchie_controlling_2015}
Daniel Ritchie, Ben Mildenhall, Noah~D. Goodman, and Pat Hanrahan.
\newblock Controlling {Procedural} {Modeling} {Programs} with {Stochastically}-ordered {Sequential} {Monte} {Carlo}.
\newblock {\em ACM Transactions on Graphics}, 34(4), 2015.
\newblock Article 105.

\bibitem[\protect\citeauthoryear{Ritchie \bgroup \em et al.\egroup }{2018}]{ritchie_example-based_2018}
Daniel Ritchie, Sarah Jobalia, and Anna Thomas.
\newblock Example-based {Authoring} of {Procedural} {Modeling} {Programs} with {Structural} and {Continuous} {Variability}.
\newblock {\em Computer Graphics Forum}, 37(2):401--413, 2018.

\bibitem[\protect\citeauthoryear{Rowe}{2016}]{rowe_inspired_2016}
Brian Rowe.
\newblock Inspired {Space}: {Inside} the {Development} of {Astroneer}, 2016.
\newblock Available at: https://www.unrealengine.com/en-US/developer-interviews/inspired-space-inside-the-development-of-astroneer.

\bibitem[\protect\citeauthoryear{Schaffrath}{1995}]{schaffrath_essen_1995}
Helmut Schaffrath.
\newblock {\em The {Essen} {Folksong} {Collection}}.
\newblock Center for Computer Assisted Research in the Humanities, 1995.

\bibitem[\protect\citeauthoryear{Schoenberg \bgroup \em et al.\egroup }{1967}]{schoenberg_fundamentals_1967}
Arnold Schoenberg, Gerald Strang, and Leonard Stein.
\newblock {\em Fundamentals of {Musical} {Composition}}.
\newblock Faber and Faber, London, 1967.

\bibitem[\protect\citeauthoryear{Steedman}{1984}]{steedman_generative_1984}
Mark~J. Steedman.
\newblock A {Generative} {Grammar} for {Jazz} {Chord} {Sequences}.
\newblock {\em Music Perception: An Interdisciplinary Journal}, 2(1):52--77, 1984.

\bibitem[\protect\citeauthoryear{Talton \bgroup \em et al.\egroup }{2012}]{talton_learning_2012}
Jerry Talton, Lingfeng Yang, Ranjitha Kumar, Maxine Lim, Noah Goodman, and Radomír Měch.
\newblock Learning {Design} {Patterns} with {Bayesian} {Grammar} {Induction}.
\newblock In {\em Proceedings of the 25th {Annual} {ACM} {Symposium} on {User} {Interface}, {Software}, and {Technology}}, pages 63--74, Cambridge, Massachusetts, USA, 2012.

\bibitem[\protect\citeauthoryear{{The Church of Jesus Christ of Latter-day Saints}}{1985}]{the_church_of_jesus_christ_of_latter-day_saints_hymns_1985}
{The Church of Jesus Christ of Latter-day Saints}.
\newblock {\em Hymns of the {Church} of {Jesus} {Christ} of {Latter}-day {Saints}}.
\newblock The Church of Jesus Christ of Latter-day Saints, Salt Lake City, 1985.

\bibitem[\protect\citeauthoryear{Wolff}{2011}]{wolff_algorithm_2011}
J~Wolff.
\newblock An {Algorithm} for the {Segmentation} of an {Artificial} {Language} {Analogue}.
\newblock {\em British Journal of Psychology}, 66:79 -- 90, 2011.

\bibitem[\protect\citeauthoryear{Zeng and Lau}{2021}]{zeng_reinforcement_2021}
Te~Zeng and Francis C.~M. Lau.
\newblock {Automatic Melody Harmonization via Reinforcement Learning by Exploring Structured Representations for Melody Sequences}.
\newblock {\em Electronics}, 10(20), 2021.
\newblock Article 2469.

\bibitem[\protect\citeauthoryear{Ziv and Lempel}{1978}]{ziv_universal_1978}
Jacob Ziv and Abraham Lempel.
\newblock Compression of {Individual} {Sequences} via {Variable-rate} {Coding}.
\newblock {\em IEEE Transactions on Information Theory}, 24(5):530--536, 1978.

\bibitem[\protect\citeauthoryear{Št'ava \bgroup \em et al.\egroup }{2010}]{stava_inverse_2010}
O.~Št'ava, B.~Beneš, R.~Měch, D.~G. Aliaga, and P.~Krištof.
\newblock Inverse {Procedural} {Modeling} by {Automatic} {Generation} of {L}-systems.
\newblock {\em Computer Graphics Forum}, 29(2):665--674, 2010.

\end{thebibliography}

\section*{Appendix}\label{app:appendix}
\begin{appendix}

\section{Background and Related Work}\label{chp:chapter2}
Techniques for segmenting musical sequences into labeled components can roughly be divided into four algorithmic families: clustering algorithms that operate on self-similarity matrices, multidimensional pattern matching techniques, optimization via deep neural networks, and grammatical induction. We begin with a formal description of the musical phrase segmentation task and then describe previous work related to each of the areas mentioned above.

\subsection{Problem Formulation}\label{sec:probform}
The musical phrase segmentation task involves assigning identifying labels to contiguous regions of a musical sequence where each region delimits a musical phrase. We define a musical phrase to be a short passage of music that forms a distinct musical thought \cite{green_tonal_1965}. If similar phrases are instances of a repeated pattern, then they will share the same identifying label. The task of musical phrase segmentation can be stated more formally as follows: Given an input sequence of tokens $\omega$, output a set of discovered phrases $Q = \{ q_1, q_2,\dots \}$ such that $|Q| \leq |\omega|$ and each pattern $q_i \in Q$ is a set of occurrences $O$. An occurrence $o \in O$ is pair of integers that define the starting and ending indices of a contiguous subsequence in $\omega$. Identifying labels can be created manually or automatically by enumerating $Q$ and assigning each $q_i \in Q$ the label $i$.

Our definition for the musical phrase segmentation task is related to two similar tasks sponsored by the Music and Information Retrieval Exchange (MIREX) organization. The first task is called the \emph{structural segmentation task}, and is defined as follows: Given a musical sequence $\omega$, partition $\omega$ into phrases with identifying labels such that similar phrases receive the same label and the entire sequence is covered.\footnote{\url{https://www.music-ir.org/mirex/wiki/2017:Structural_Segmentation}} The second task, known as the \emph{discovery of repeated themes and sections task}, poses a similar problem: Given a musical sequence $\omega$, produce  a set of repeated patterns, including overlapping and nested patterns.\footnote{\url{https://www.music-ir.org/mirex/wiki/2017:Discovery_of_Repeated_Themes_\%26_Sections} } The musical phrase segmentation task defined above can be thought of as a hybrid of these two MIREX tasks. Like the \emph{structural segmentation task}, the musical phrase segmentation task requires an algorithm to partition $\omega$ into phrases with identifying labels; however, these phrase partitions need not cover the entire sequence. Additionally, we do not require phrases to be instances of a repeated pattern nor do we allow for overlapping or nested patterns, both of which are features of the \emph{discovery of repeated themes and sections task}. This new formulation captures the intuition that a musical sequence can contain singleton  phrases in addition to repeated patterns and, furthermore, that not all subsequences are musically interesting enough to constitute a phrase --- phrase segmentation algorithms should identify salient phrases without necessarily partitioning the entire sequence into labeled chunks. Additionally, an overwhelming amount of ground truth annotation data consists of disjoint rather than overlapping or nested phrase boundaries, thus we constrain the labels assigned by a musical segmentation algorithm to cover disjoint regions.

\subsection{Clustering Algorithms Using Self-Similarity Matrices}

One popular approach to phrase segmentation involves the use of the self-similarity matrices. A self-similarity matrix is a matrix $M$ computed from a sequence $\omega^\Phi$ of feature vectors  of length $n$ such that 

\begin{equation}
\begin{array}{cc}
       M_{i, j} = d(\omega^\Phi_i, \omega^\Phi_j), & i, j \in (1,...,n)
\end{array}
\end{equation}

\noindent where the function $d$ computes the distance between individual elements (feature vectors) of $\omega^\Phi$ (here we are using the subscript $\omega^\Phi_i$ to denote the $i$-th elements of $\omega^\Phi$). Repeated patterns in a sequence will appear as contiguous diagonal ``stripes'' in the upper triangular portion of $M$ and are discovered using a clustering algorithm. \cite{nieto_identifying_2014} propose \textsc{MotivesExtractor}, an algorithm which uses chromagraphs of audio sequences to generate $M$. Pattern occurrences are grouped together by first choosing a canonical stripe based on length and distance from the diagonal and then scanning the remaining columns for stripes that match the canonical stripe against a threshold. Although originally intended to work with audio data, the clustering technique can be converted to work with symbolic music sequences by constructing $M$ from a sequence of appropriate feature vectors derived from a symbolic musical sequence. \cite{bodily_inferring_2021} improve on this technique by  parameterizing the distance function $d$ with a vector of weights $\boldsymbol{\theta}$ that is learned via a genetic algorithm. Pattern annotations are produced by using a threshold to locate several maximal points in $M$ and then backtracking along paths produced by the Smith-Waterman algorithm to find pattern occurrences.

\subsection{Multidimensional Pattern Matching}

Meredith et. al. were the first to apply point-set algorithms to the problem of musical sequence segmentation with the introduction of the \textsc{SIATEC} algorithm \cite{meredith_algorithms_2002}, which works as follows. First, a symbolic musical sequence $\omega$ is converted into a sequence of feature vectors $\omega^\Phi$. Next, a translation vector is computed for every pair of feature vectors in the dataset via vector subtraction. Minuends from the previous step that produced the same translation vector are grouped together into a set. Each of these sets, which now comprises a subset of the original dataset, are termed Maximally Translatable Patterns (MTPs) and interpreted as patterns in the original musical sequence. Occurrences for each pattern, termed Translation Equivalence Classes (TECs), are derived by grouping all vectors that map to the same MTP.  \textsc{SIATEC} was designed to handle both monophonic and polyphonic musical sequences with feature vectors of arbitrary dimension. Further iterations of this approach can be found in \cite{meredith_cosiatec_2013,meredith_music_2015,louboutin_using_2016}.

\subsection{Machine Learning}

There are many approaches to generating musical sequences using deep neural networks, but applying deep learning techniques specifically to the task of partitioning a musical sequence into labeled sections is a relatively nascent innovation. In the audio domain, convolutional neural networks (CNNs) are capable of producing musical phrase boundaries when given spectrograms of musical sequences \cite{olivan_boundary_2020}. In \cite{zeng_reinforcement_2021}, the authors cast the phrase segmentation problem in terms of a reinforcement learning task where an agent is given a state --- represented as a note event $e_i$ --- and decides whether or not $e_i$ is a phrase boundary according to a learned policy $\pi$.  \cite{guan2018melodic} perform a comparison of different types of deep learning architectures, including CNNs and Long-Short Term Memory networks, on symbolic musical data and find that CNNs combined with conditional random field techniques yield the best results. Other efforts in this algorithm family include the use of Restricted Boltzmann Machines \cite{lattner_probabilistic_2015}. 

\subsection{Grammatical Induction}\label{sec:gI}

When properly interpreted, context-free grammars provide a viable solution to the musical phrase segmentation problem because they have the ability to organize repeated patterns into compact, labeled representations. Unlike the methods described above, they also have the potential to unveil the hierarchical nature of musical sequences.   A context-free grammar $G$ is formally defined as a tuple $(\pazocal{N}, \pazocal{T}, \pazocal{R}, N_s)$ where $\pazocal{N}$ is a set of \textit{non-terminal} symbols and $\pazocal{T}$ is the set of \textit{terminal} symbols such that $\pazocal{N} \cap \pazocal{T} = \varnothing$. Additionally, $\pazocal{R}$ is a set of transformation \textit{rules} (also called \textit{productions}) that define symbol substitutions of the form $\pazocal{N} \rightarrow{\{\pazocal{T} \cup \pazocal{N} \}^*}$ and $N_s \in \pazocal{N}$ is a non-terminal symbol that is designated as the start symbol. Given two sequences of symbols\footnote{Unless otherwise noted, we use the term \textit{symbol} to refer to both terminal and non-terminal symbols} $\alpha,\ \beta \in \{ \pazocal{T} \cup \pazocal{N} \}^*$, $\alpha$ is said to be derived from $\beta$ if and only if a finite number of applications of $\pazocal{R}$ to $\alpha$ can produce $\beta$, denoted by $\alpha \xrightarrow[]{\pazocal{R}} \beta$. The language $L_G$ generated by a grammar $G$ is defined as $\{ \omega \in \pazocal{T}^* \mid N_s \xrightarrow[]{\pazocal{R}} \omega \}$. When $\omega$ is a musical sequence, the non-terminals in a context-free grammar $G$ can be interpreted as identifiers of musical phrases --- each appearance of a non-terminal $N_i$ in the right-hand side of $G$'s start rule corresponds to an occurrence of the phrase $\{ \omega_i \in \pazocal{T}^* \mid N_i \xrightarrow[]{\pazocal{R}} \omega_i \}$ with $N_i$ serving as the phrase label. We can obtain $G$ via the process of grammatical induction. 

A grammatical induction algorithm $f : \pazocal{T}^* \rightarrow{\pazocal{G}}$ infers a context-free grammar $G \in \pazocal{G}$ from an input sequence $\omega \in \pazocal{T}^*$ such that $f(\omega) = G$. We focus specifically on ``lossless'', non-probabilistic grammatical induction algorithms that add the additional stipulation $L_G = \{ \omega \}$ --- that is, the language generated by the resulting grammar is precisely the original input sequence $\omega$ --- due to their efficacy as data compression algorithms. In the literature, approaches to this kind of grammatical induction algorithm fall into two categories: online algorithms that process $\omega$ one token at a time while producing a new grammar $G$ at each step \cite{nevill-manning_identifying_1997,ziv_universal_1978}, and offline algorithms which consume the entirety of $\omega$ upfront and then selectively create production rules that optimize $G$ according to some objective function \cite{larsson_offline_1999,wolff_algorithm_2011,nevill-manning_-line_2000}. Traditionally, these algorithms operate on sequences of strings, individual characters, or bytes, but we employ generalized implementations that are able to operate on generic sequences. In our experiments we run the grammatical induction algorithms on sequences of feature vectors created by musical viewpoint combinations \cite{conklin_feature_2008}.

This study would not mark the first time that grammatical induction algorithms have been used to identify musical phrases. \cite{kitani_improvgenerator:_2010}, for example, encode cajon performances as  sequences of duration values and then feed the resulting sequences into \sequitur{} \cite{nevill-manning_identifying_1997} to infer musical phrases. It is unclear, however, as to why they chose that particular algorithm and viewpoint; it is possible that a different algorithm and choice of viewpoints may have produced better results. In addition to \sequitur{}, we evaluate four other grammatical induction algorithms and 31 unique viewpoint combinations to see which produce the best labeled sequences when compared to human annotations. The induction algorithms we study are well-known for their data compression attributes, and we are hoping to discover that their success in one domain can yield viable solutions for the musical phrase segmentation task.

\clearpage

\begin{figure*}[t]
    \centering
    \begin{subfigure}[]{\linewidth}
        \includegraphics[width=\linewidth,keepaspectratio]{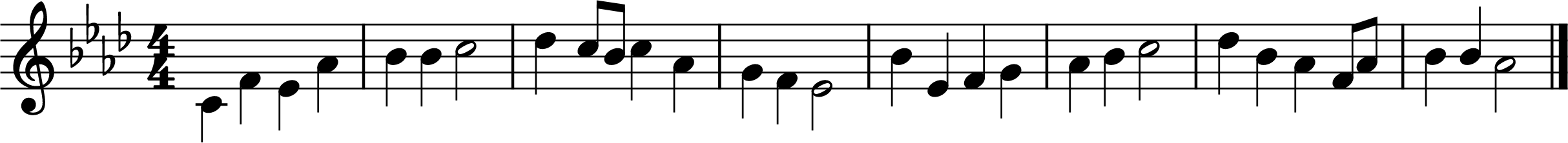}
        \caption{Visual representation}
    \end{subfigure}\\
    \vspace{.5cm}
    \begin{subfigure}[]{\linewidth}
        \tiny
        \centering
        \setlength\tabcolsep{1.3pt}
        \begin{tabular}{lllllllllllllllllllllllllllllll}
        \hline
        \multicolumn{1}{|l}{Index} & 0   & 1   & 2   & 3   & 4   & 5   & 6   & 7   & 8   & 9   & 10   & 11   & 12   & 13   & 14   & 15   & 16   & 17   & 18   & 19   & 20   & 21   & 22   & 23   & 24   & 25   & 26   & 27   & 28   & \multicolumn{1}{l|}{29} \\ \hline
        Onset                      & 0.0 & 1.0 & 2.0 & 3.0 & 4.0 & 5.0 & 6.0 & 8.0 & 9.0 & 9.5 & 10.0 & 11.0 & 12.0 & 13.0 & 14.0 & 16.0 & 17.0 & 18.0 & 19.0 & 20.0 & 21.0 & 22.0 & 24.0 & 25.0 & 26.0 & 27.0 & 27.5 & 28.0 & 29.0 & 30.0                    \\
        Pitch                      & 60  & 65  & 63  & 68  & 70  & 70  & 72  & 73  & 72  & 70  & 72   & 68   & 67   & 65   & 63   & 70   & 63   & 65   & 67   & 68   & 70   & 72   & 73   & 70   & 68   & 65   & 68   & 70   & 70   & 68                      \\
        Duration                   & 1.0 & 1.0 & 1.0 & 1.0 & 1.0 & 1.0 & 2.0 & 1.0 & 0.5 & 0.5 & 1.0  & 1.0  & 1.0  & 1.0  & 2.0  & 1.0  & 1.0  & 1.0  & 1.0  & 1.0  & 1.0  & 2.0  & 1.0  & 1.0  & 1.0  & 0.5  & 0.5  & 1.0  & 1.0  & 2.0                    
        \end{tabular}
        \caption{Note event representation}
    \end{subfigure}
    \caption[Representation of the note event sequence $\omega^\pazocal{E}$ for Hymn 5]{Representation of the note event sequence $\omega^\pazocal{E}$ for Hymn 5. (a) shows a visual rendering of the musical sequence and (b) shows the note event representation. A column in (b) shows the note event $e_i$ that is equivalent to the 3-tuple $(o, p, d)$ where $o$, $p$ and $d$ represent the onset time, the midi pitch value, and  duration (in quarter notes). A note event sequence $\omega^\pazocal{E}$ is comprised of note events $e_1, e_2, ..., e_n$.}
    \label{fig:noteevents}
\end{figure*}

\begin{figure*}[t]
    \centering
    \begin{subfigure}[c]{.3\linewidth}
        \footnotesize
        \begin{tabular}{c}
\begin{lstlisting}
Phrase A:
    Occ 1: [0,7]
    Occ 2: [53,62]
Phrase B:
    Occ 1: [8,17]
    Occ 2: [26,35]
Phrase C:
    Occ 1: [18,25]
Phrase D:
    Occ 1: [36,42]
Phrase E:
    Occ 1: [43,52]
Phrase F:
    Occ 1: [63,73]
\end{lstlisting}
        \end{tabular}
        \caption{Textual representation}
    \end{subfigure}%
    \begin{subfigure}[c]{.7\linewidth}
        \includegraphics[width=\linewidth,keepaspectratio]{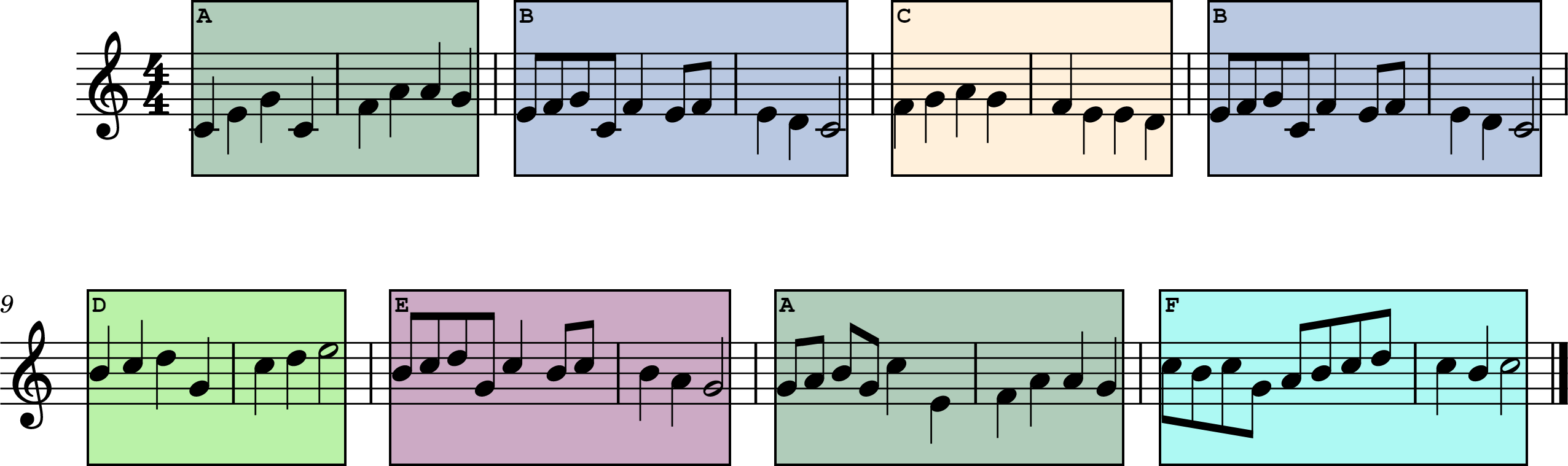}
        \caption{Visual representation}
    \end{subfigure}
    \caption[Ground truth annotation for Hymn 2]{Ground truth phrase annotation $P$ for Hymn 2 given by Annotator 1 represented textually (a) and visually (b). Each phrase $p \in P$ was given an identifying label and contains one or more occurrences. Each occurrence $o$ of the phrase $p_i$ is equivalent to the range $[x_1,\ x_2]$ such that $x_1 < x_2 \in \mathbb{N}$, where $x_1$ and $x_2$ represent the starting and ending indices of that particular phrase occurrence. In (b), each distinct notehead represents a note event $e_i$ at index $i$ (0-based). Phrases that contain 2 or more occurrences are known as patterns --- in this example, Phrases A and B are considered patterns.}
    \label{fig:ann1example}
\end{figure*}

\begin{figure*}[t]
    \centering
    \begin{subfigure}[]{\linewidth}
        \includegraphics[width=\linewidth,keepaspectratio]{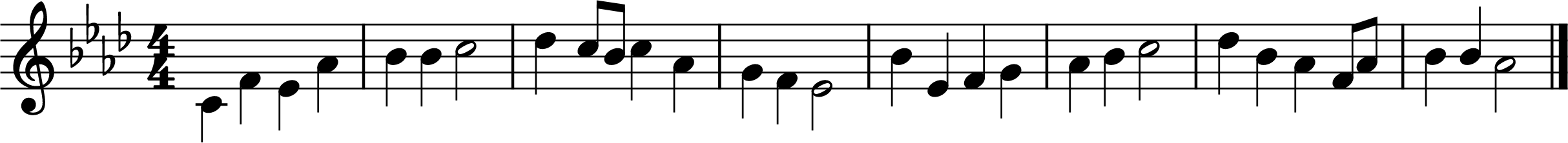}
        \caption{Visual representation}
    \end{subfigure}\\
    \vspace{.5cm}
    \begin{subfigure}[]{\linewidth}
        \tiny
        \centering
        \setlength\tabcolsep{1.3pt}
        \begin{tabular}{lllllllllllllllllllllllllllllll}
        \hline
        \multicolumn{1}{|l}{Index} & 0   & 1   & 2   & 3   & 4   & 5   & 6   & 7   & 8   & 9   & 10   & 11   & 12   & 13   & 14   & 15   & 16   & 17   & 18   & 19   & 20   & 21   & 22   & 23   & 24   & 25   & 26   & 27   & 28   & \multicolumn{1}{l|}{29} \\ \hline
        Onset                      & 0.0 & 1.0 & 2.0 & 3.0 & 4.0 & 5.0 & 6.0 & 8.0 & 9.0 & 9.5 & 10.0 & 11.0 & 12.0 & 13.0 & 14.0 & 16.0 & 17.0 & 18.0 & 19.0 & 20.0 & 21.0 & 22.0 & 24.0 & 25.0 & 26.0 & 27.0 & 27.5 & 28.0 & 29.0 & 30.0                    \\
        Pitch                      & 60  & 65  & 63  & 68  & 70  & 70  & 72  & 73  & 72  & 70  & 72   & 68   & 67   & 65   & 63   & 70   & 63   & 65   & 67   & 68   & 70   & 72   & 73   & 70   & 68   & 65   & 68   & 70   & 70   & 68                      \\
        Duration                   & 1.0 & 1.0 & 1.0 & 1.0 & 1.0 & 1.0 & 2.0 & 1.0 & 0.5 & 0.5 & 1.0  & 1.0  & 1.0  & 1.0  & 2.0  & 1.0  & 1.0  & 1.0  & 1.0  & 1.0  & 1.0  & 2.0  & 1.0  & 1.0  & 1.0  & 0.5  & 0.5  & 1.0  & 1.0  & 2.0                    
        \end{tabular}
        \caption{Note event representation}
    \end{subfigure}\\
    \vspace{.5cm}
    \begin{subfigure}[]{\linewidth}
        \tiny
        \centering
        \setlength\tabcolsep{2.5pt}
        \begin{tabular}{lllllllllllllllllllllllllllllll}
            \hline
            \multicolumn{1}{|l}{Index} & 0      & 1   & 2   & 3   & 4   & 5   & 6   & 7   & 8   & 9   & 10  & 11  & 12  & 13  & 14  & 15  & 16  & 17  & 18  & 19  & 20  & 21  & 22  & 23  & 24  & 25  & 26  & 27  & 28  & \multicolumn{1}{l|}{29} \\ \hline
            $\fn{pitch}$               & 60     & 65  & 63  & 68  & 70  & 70  & 72  & 73  & 72  & 70  & 72  & 68  & 67  & 65  & 63  & 70  & 63  & 65  & 67  & 68  & 70  & 72  & 73  & 70  & 68  & 65  & 68  & 70  & 70  & 68                      \\
            $\fn{duration}$            & 1.0    & 1.0 & 1.0 & 1.0 & 1.0 & 1.0 & 2.0 & 1.0 & 0.5 & 0.5 & 1.0 & 1.0 & 1.0 & 1.0 & 2.0 & 1.0 & 1.0 & 1.0 & 1.0 & 1.0 & 1.0 & 2.0 & 1.0 & 1.0 & 1.0 & 0.5 & 0.5 & 1.0 & 1.0 & 2.0                     \\
            $\fn{ioi}$                 & $\bot$ & 1.0 & 1.0 & 1.0 & 1.0 & 1.0 & 1.0 & 2.0 & 1.0 & 0.5 & 0.5 & 1.0 & 1.0 & 1.0 & 1.0 & 2.0 & 1.0 & 1.0 & 1.0 & 1.0 & 1.0 & 1.0 & 2.0 & 1.0 & 1.0 & 1.0 & 0.5 & 0.5 & 1.0 & 1.0                     \\
            $\fn{pitchC}$              & $\bot$ & 1   & -1  & 1   & 1   & 0   & 1   & 1   & -1  & -1  & 1   & -1  & -1  & -1  & -1  & 1   & -1  & 1   & 1   & 1   & 1   & 1   & 1   & -1  & -1  & -1  & 1   & 1   & 0   & -1                      \\
            $\fn{pitchI}$              & $\bot$ & 5   & -2  & 5   & 2   & 0   & 2   & 1   & -1  & -2  & 2   & -4  & -1  & -2  & -2  & 7   & -7  & 2   & 2   & 1   & 2   & 2   & 1   & -3  & -2  & -3  & 3   & 2   & 0   & -2                     
        \end{tabular}
        \caption{Viewpoint representation (VCI-31)}
    \end{subfigure}
    \caption[Representation of VCI-31 for Hymn 5]{Representation of VCI-31, which is equal to \texttt{\{ pitch, duration, ioi, pitchC, pitchI \}}, for Hymn 5. The top (a) and middle (b) figures are repeated from Figure~\ref{fig:noteevents}. A column in (c) shows each element of the feature vector $\phi_i$, where $i$ is indicated by the index. For example, $\phi_0 = \langle 60, 1.0, \bot, \bot, \bot \rangle$ and $\phi_9 = \langle 70, 0.5, 0.5, -1, -2 \rangle$. An algorithm is given an entire sequence of feature vectors $\omega^\Phi = \phi_0, \phi_1, ..., \phi_n$ as input.}
    \label{fig:vpexample}
\end{figure*}

\begin{figure*}
  \centering
    \includegraphics[width=0.9\textwidth,height=0.9\textheight,keepaspectratio]{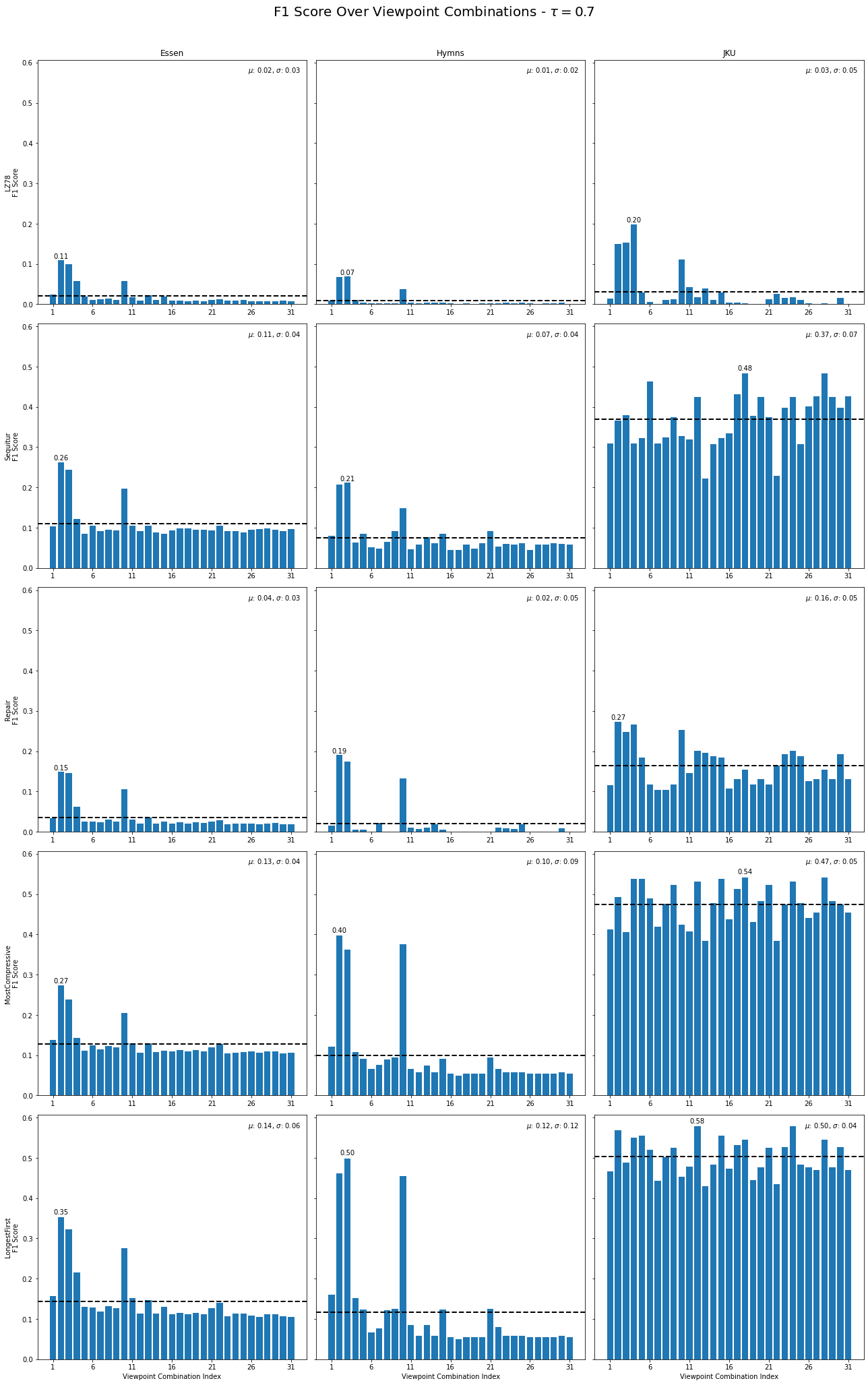}
  \caption[Summary of F1 scores for each algorithm and dataset]{Summary of F1 scores for each algorithm and dataset.  Each individual plot shows the average F1 score (y axis) obtained by a given algorithm on a given dataset when using each viewpoint combination (x axis). Each plot is annotated with the mean --- denoted as $\mu$ and the dashed black line --- and variance (denoted as $\sigma$) of all F1 scores across all viewpoints for that particular algorithm and dataset combination. The columns of the figure correspond to the dataset used, and the rows correspond to the algorithm used. }
  \label{fig:f1Results}
\end{figure*}

\begin{figure*}[t]
  \centering
    \includegraphics[width=\linewidth,keepaspectratio]{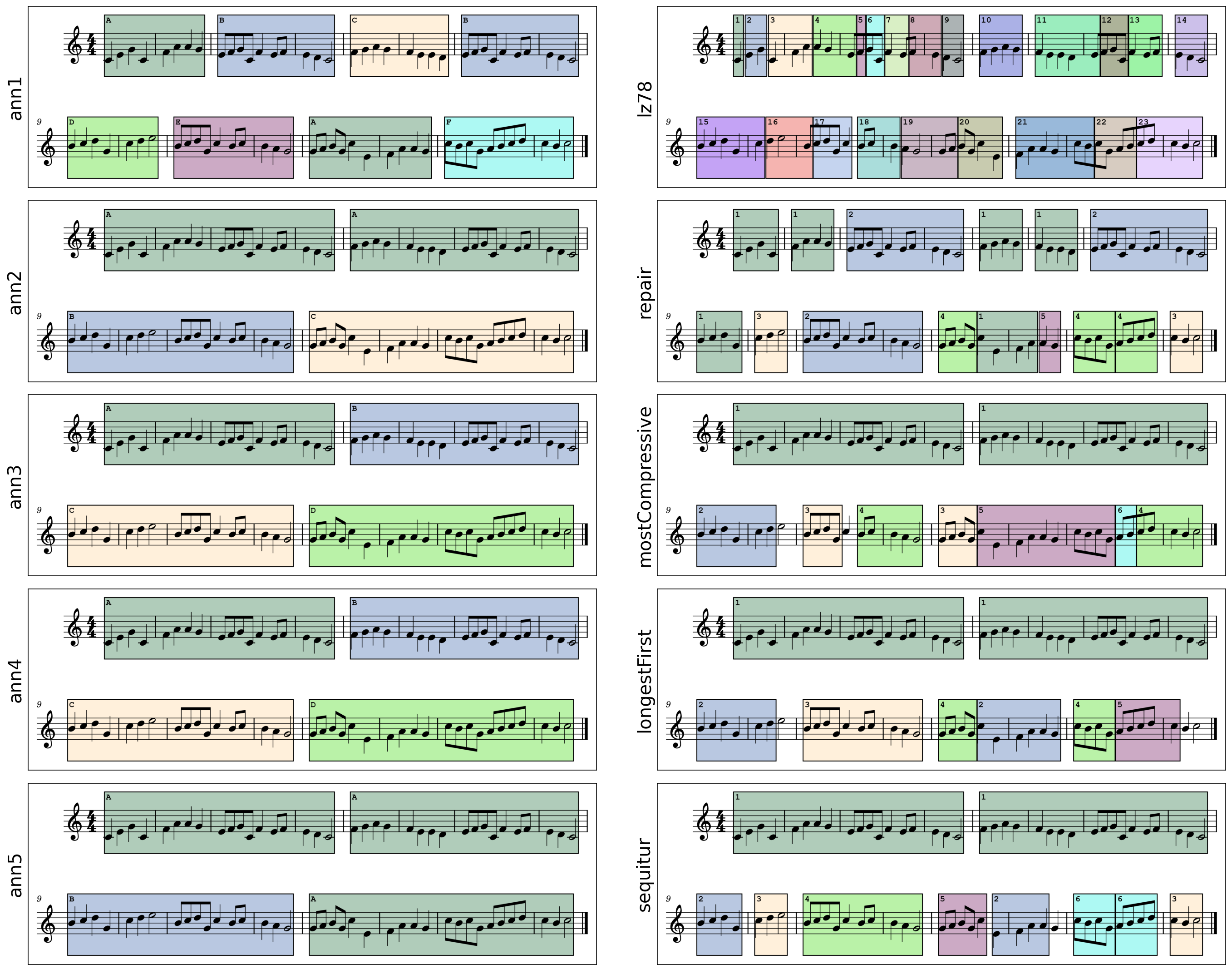}
  \caption[Ground truth and discovered annotations for Hymn 2]{Ground truth and discovered annotations for Hymn 2. The first column contains annotations from five different annotators and the second column contains annotations generated by each grammatical induction algorithm.}
  \label{fig:hymn2}
\end{figure*}

\begin{figure}[t]
    \centering
    \includegraphics[width=3in,keepaspectratio]{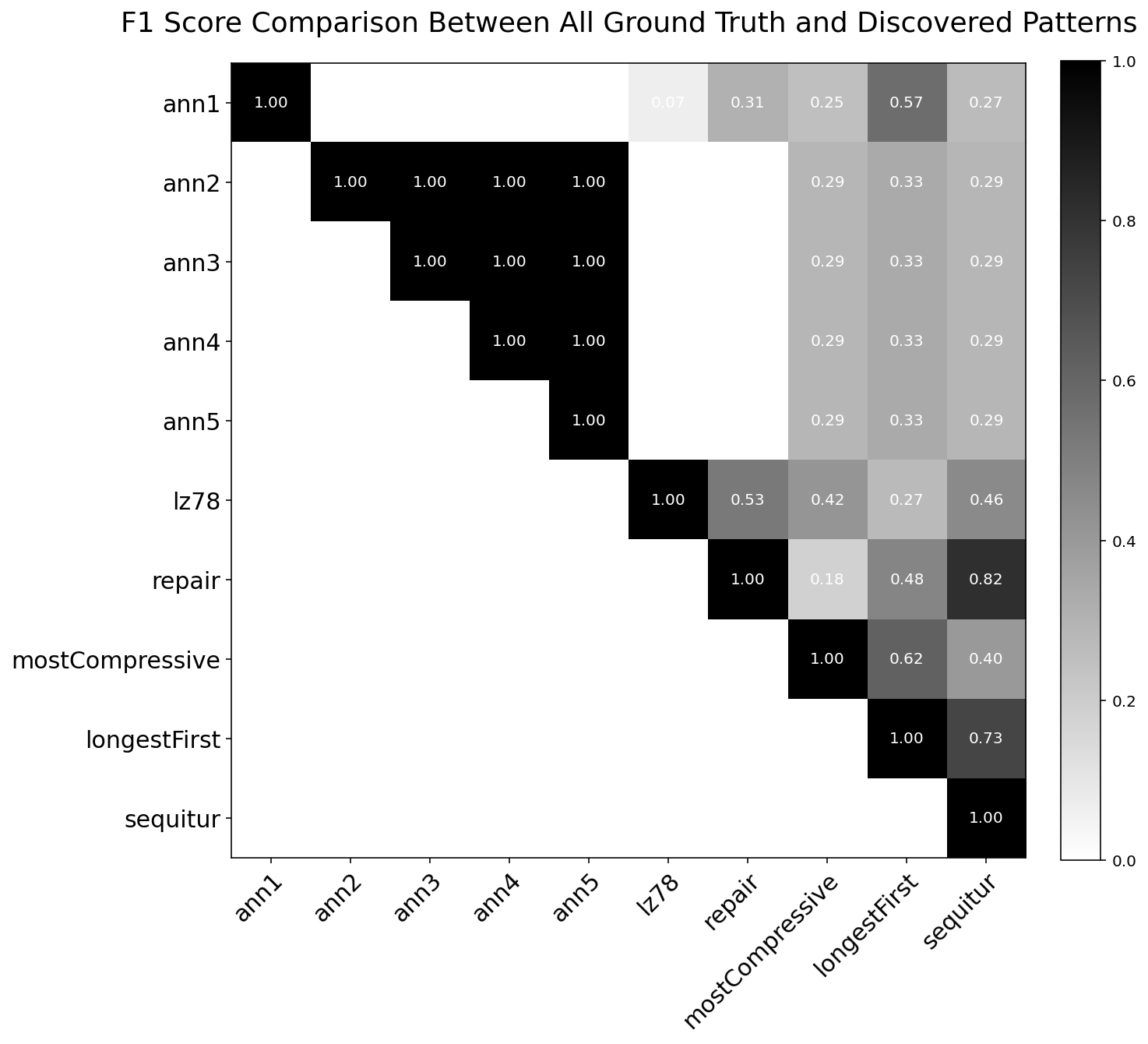}
    \caption[Heat map showing the F1 scores for Hymn 2]{Heat map showing the F1 scores generated using \textsc{FuzzyIntersection} ($\tau$ = 0.7) for all ground truth and discovered patterns of Hymn 2 using the \fn{duration} viewpoint, which corresponds with the annotations shown in Figure~\ref{fig:hymn2}. Scores below the main diagonal are not shown.}
    \label{fig:f1Comp}
\end{figure}

\begin{figure*}[t]
    \centering
    \includegraphics[width=\linewidth,keepaspectratio]{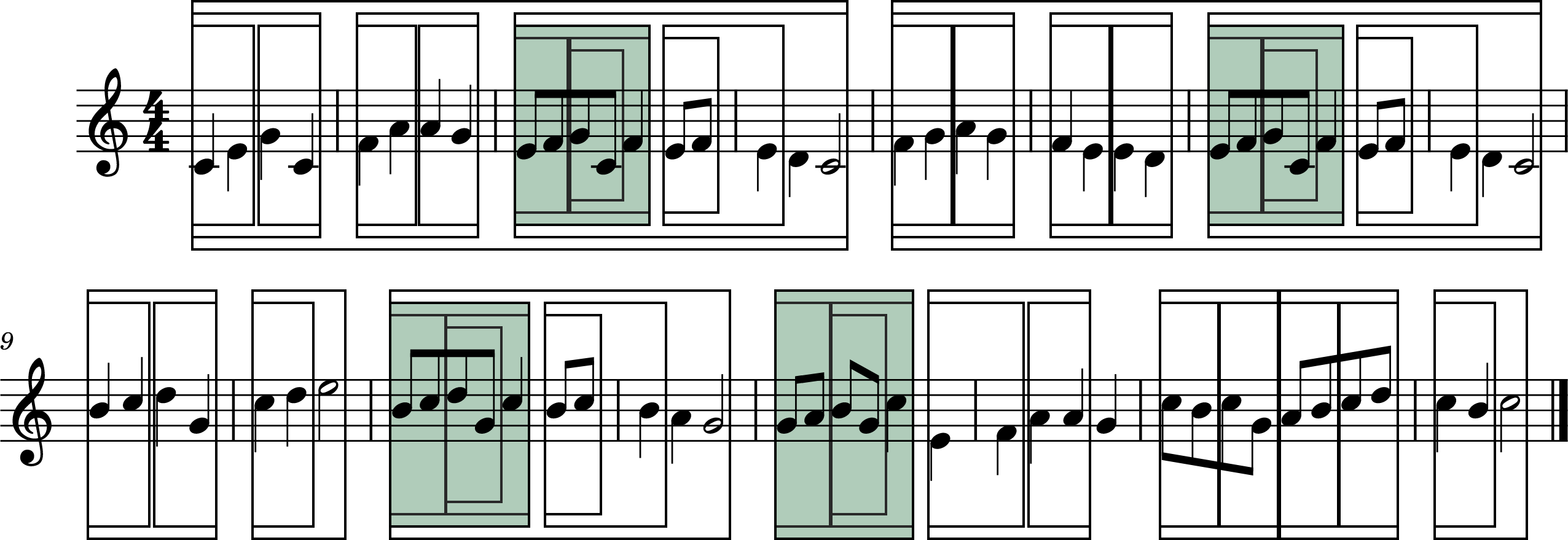}
    \caption[A visual representation of the grammar generated for Hymn 2 by \sequitur{} using the \fn{duration} viewpoint]{A visual representation of the grammar shown in Figure~\ref{fig:gram} in the main text generated for Hymn 2 by \sequitur{} using the \fn{duration} viewpoint, with rule names omitted for visual clarity. Each ``box'' represents a non-terminal in the grammar (the box which represents $N_s$ is not shown). For example, the boxes that represent rule $N_1$ are the outermost boxes that span bars 1--4 and 5--8 (these are exactly the same boxes that correspond to \sequitur{}'s Pattern 1 in Figure~\ref{fig:hymn2}, because Pattern 1 was extracted from $N_1$). They each contain (as direct descendants) three nested boxes: two for rule $N_2$ and one for rule $N_4$. The occurrences of rule $N_5$ are highlighted as a visual reference. Because we are only using the outermost boxes to designate phrase boundaries, the nested occurrences of $N_5$ are subsumed in the annotation process.}
    \label{fig:hier}
\end{figure*}

\end{appendix}

\end{document}